\documentclass{article}
\usepackage{ijcai26}
\usepackage{stfloats}
\usepackage{enumitem}
\setlist[itemize]{topsep=2pt,itemsep=1pt,parsep=0pt,partopsep=0pt,leftmargin=*}
\setlist[enumerate]{topsep=2pt,itemsep=1pt,parsep=0pt,partopsep=0pt,leftmargin=*}

\usepackage{titlesec}

\titlespacing*{\section}{0pt}{6pt plus 2pt minus 2pt}{4pt plus 2pt minus 2pt}
\titlespacing*{\subsection}{0pt}{5pt plus 2pt minus 2pt}{3pt plus 1pt minus 1pt}
\titlespacing*{\subsubsection}{0pt}{4pt plus 2pt minus 2pt}{2pt plus 1pt minus 1pt}
\titlespacing*{\paragraph}{0pt}{3pt plus 1pt minus 1pt}{6pt}

\usepackage{times}
\usepackage{soul}
\usepackage{url}
\usepackage[hidelinks]{hyperref}
\usepackage[utf8]{inputenc}
\usepackage[small]{caption}
\usepackage{graphicx}
\usepackage{amsmath}
\usepackage{amsthm}
\usepackage{booktabs}
\usepackage{algorithm}
\usepackage{algorithmic}
\usepackage[switch]{lineno}

\title{Harmonizing the Deep: A Unified Information Pipeline for Robust Marine Biodiversity Assessment Across Heterogeneous Domains}

\author{
    Author Name
    \affiliations
    Affiliation
    \emails
    email@example.com
}

\author{
Marco Piccolo$^1$
\and
Qiwei Han$^2$\and
Astrid van Toor$^3$\and
Joachim Vanneste$^4$\and
\affiliations
$^{1,2}$Nova School of Business \& Economics, Carcavelos Portgual\\
$^{3,4}$blueOASIS, Ericeira, Portugal\\
\emails
\{63996, qiwei.han\} @novasbe.pt,
\{avtoor, jvanneste\}@blueoasis.pt,
}

\usepackage{tikz}
\usetikzlibrary{arrows.meta,positioning,fit}
\usepackage{helvet}
\usepackage{caption}
\usepackage{graphicx} 

\definecolor{LineCol}{HTML}{000000}
\definecolor{TextCol}{HTML}{000000}

\tikzset{
  box/.style={
    draw=LineCol, fill=white,
    rounded corners=2pt,
    align=center, inner sep=4pt,
    minimum width=64mm, minimum height=9mm,
    text=TextCol, line width=0.5pt
  },
  boxAlt/.style={
    draw=LineCol, fill=white,
    rounded corners=2pt,
    align=center, inner sep=4pt,
    minimum width=64mm, minimum height=9mm,
    text=TextCol, line width=0.5pt
  },
  datasetBox/.style={
    draw=LineCol, fill=white,
    rounded corners=2pt,
    align=center, inner sep=2pt,
    minimum width=22mm,
    text=TextCol, line width=0.45pt
  },
  group/.style={
    draw=LineCol, rounded corners=2pt,
    inner sep=3pt, line width=0.45pt
  },
  note/.style={
    draw=LineCol, fill=white,
    rounded corners=2pt,
    align=left, inner sep=3pt,
    minimum width=40mm,
    text=TextCol, line width=0.4pt
  },
  arrow/.style={-Latex, draw=LineCol, line width=0.55pt},
  dashedarrow/.style={-Latex, draw=LineCol, dashed, line width=0.4pt},
  title/.style={font=\bfseries, text=TextCol}
}

\newcommand{\bb}{\raisebox{0.2ex}{\tiny$\bullet$}\ }

\begin{document}

\maketitle

\begin{abstract} Marine biodiversity monitoring requires scalability and reliability across complex underwater environments to support conservation and invasive-species management. Yet existing detection solutions often exhibit a pronounced deployment gap, with performance degrading sharply when transferred to new sites. This work establishes the foundational detection layer for a multi-year invasive species monitoring initiative targeting Arctic and Atlantic marine ecosystems. We address this challenge by developing a Unified Information Pipeline that standardises heterogeneous datasets into a comparable information flow and evaluates a fixed, deployment-relevant detector under controlled cross-domain protocols. Across multiple domains, we find that \emph{structural} factors, such as scene composition, object density, and contextual redundancy, explain cross-domain performance loss more strongly than \emph{visual} degradation such as turbidity, with sparse scenes inducing a characteristic \emph{``Context Collapse''} failure mode. We further validate operational feasibility by benchmarking inference on low-cost edge hardware, showing that runtime optimisation enables practical sampling rates for remote monitoring. The results shift emphasis from image enhancement toward structure-aware reliability, providing a democratised tool for consistent marine ecosystem assessment. \end{abstract}

\section{Introduction}

Marine ecosystem monitoring provides essential evidence for understanding ocean health and supporting timely management responses to environmental change~\cite{Mieszkowska2014}. Climate-driven shifts in species distributions, habitat fragmentation, and rising pressure from biological invasions threaten marine sustainability and accelerate biodiversity loss~\cite{EC2013}. In Europe, invasive species affect a substantial share of threatened taxa and continue to pose persistent risks to marine and coastal ecosystems~\cite{EUBiodiversityStrategy2020,NOAA_MarineSpecies}. These challenges align directly with UN Sustainable Development Goal 14 (Life Below Water)~\cite{UNSDG14} and the EU Biodiversity Strategy for 2030~\cite{EUBiodiversityStrategy2020}, motivating scalable, field-deployable tools that can monitor biodiversity consistently across space and time.

Conventional monitoring approaches remain costly and difficult to scale. Fisheries surveys sample only portions of populations and face accuracy and logistics constraints~\cite{Labrosse2002}. Manual visual census depends on human judgement and introduces observer bias, while environmental DNA (eDNA) requires repeated physical sampling to track diversity and abundance~\cite{rourke2022environmental,ruppert2019past}. These limitations have accelerated interest in non-destructive automated monitoring~\cite{mcgeady2023review}. In terrestrial settings, deep learning can replace time-intensive manual surveys while producing competitive population estimates~\cite{torney2019}. In underwater settings, automated fish detection can outperform experts and citizen scientists on curated benchmarks~\cite{ditria2020}. However, such results also highlight a deployment-relevant asymmetry: even when average accuracy is high, false negatives can remain elevated in operational streams, which is consequential for early-warning monitoring of rare or invasive individuals.

The central obstacle for marine deployment is therefore not achieving high accuracy on a single curated dataset, but ensuring that detection outputs remain \emph{comparable and trustworthy across deployments}~\cite{wang2022generalizing}. In practice, models tuned for one site often degrade sharply when transferred to new habitats, camera viewpoints, or ecological contexts~\cite{recht2019imagenet}. This \emph{deployment gap} forces practitioners to build ad hoc site-specific models without clear guidance on whether failures are driven primarily by visual degradation (turbidity, colour attenuation, blur) or by structural properties of the scene (object density, overlap/occlusion, contextual redundancy). For biodiversity monitoring, this distinction is operationally decisive: unstable sensitivity across sites can create spurious trends, while missed detections can suppress early warning signals.

We address this gap by reframing the underwater fish detection task~\cite{elmezain2025advancing,gonzalez2023survey,jian2024underwater,xu2023systematic} as an \emph{information reliability} problem: the goal is to produce detection outputs that remain interpretable as monitoring evidence under domain shift, subject to operational constraints. The study is co-designed with two stakeholder roles: marine ecologists, who require cross-site comparability and recall-sensitive outputs for early warning, and technical operators, who impose compute, memory, and connectivity constraints that bound feasible deployment. We develop a unified information pipeline that standardises heterogeneous datasets into a comparable representation and evaluate a fixed, deployment-relevant detector under controlled cross-domain protocols. Holding the detector fixed isolates domain effects attributable to data and environment, enabling an auditable diagnosis of when and why reliability fails under transfer.

A key methodological point is that ``scene structure'' induces two qualitatively different failure regimes. In sparse, low-redundancy scenes, isolated targets provide few contextual cues for verification, leading to a recall collapse (\emph{``Context Collapse''})~\cite{geirhos2020shortcut}. In dense, high-overlap scenes, occlusion and silhouette merging reduce separability and shift errors toward missed detections and localization failures. This two-regime framing reconciles cases where increasing density improves recall (sparse-to-moderate transition) with cases where increasing density/overlap harms recall (moderate-to-crowded transition), and it supports monitoring-facing interpretation beyond standard detection metrics (mAP, precision, recall).


Our contributions are:
\begin{itemize}
  \setlength{\itemsep}{0pt}
  \setlength{\parskip}{0pt}
  \setlength{\parsep}{0pt}
    \item \textbf{Unified reliability pipeline.} We design an end-to-end pipeline that harmonises cleaning, annotation, deduplication, and evaluation across heterogeneous underwater datasets, enabling controlled comparison of detector behaviour under monitoring-relevant constraints.
    \item \textbf{Controlled cross-domain diagnosis with covariate controls.} We introduce stress tests and image-level diagnostic covariates that separate visual degradation from scene structure, and we quantify their relative association with detection success using a mixed-effects attribution model that controls for dataset-level heterogeneity.
    \item \textbf{Mechanistic error characterization.} We complement aggregate metrics with a standard detection error decomposition (TIDE-style categories) to distinguish verification-driven misses in sparse regimes from separability-driven errors in crowded regimes~\cite{bolya2020tide}.
    \item \textbf{Deployment feasibility evidence.} We validate operational feasibility via edge benchmarking on low-cost hardware (NVIDIA Jetson Nano), showing that hardware-aware optimisation (TensorRT) supports practical sampling rates for long-horizon monitoring deployments~\cite{han2015edge}.
\end{itemize}

\paragraph{Reproducibility and artifacts.}
To support reproducibility and deployment-oriented reuse while preserving double-blind review, preprocessing scripts, metric computation code, training configurations, and benchmarking protocols are provided in anonymised supplementary material.



\section{Stakeholder Co-Design and Operational Context}
\label{sec:stakeholder}

This work follows a bottom-up co-design approach grounded in the operational realities of long-horizon underwater monitoring, as the foundational detection layer for a multi-year invasive species monitoring initiative targeting Arctic and Atlantic marine ecosystems. The study is informed by a multilateral collaboration between (i) technical operators responsible for autonomous sensing infrastructure, edge compute, and data pipelines, and (ii) marine ecologists who use derived indicators for biodiversity assessment and early-warning monitoring. To ensure the system targets deployment bottlenecks rather than abstract benchmarks, we elicited requirements through structured consultations with these stakeholder roles. These consultations directly shaped the evaluation priorities (cross-site comparability; recall-sensitive failure modes), bounded model choice by operational constraints (edge memory/latency), and defined the form of monitoring impact considered valid (comparability of indicators under domain shift rather than peak in-domain accuracy).

\paragraph{(R1) Ecologists: Cross-site comparability.}
Ecologists required that indicator trends remain comparable across sites, so that apparent changes are not artefacts of camera, viewpoint, habitat, or acquisition protocol.
\textit{Design response:} we standardise heterogeneous sources into a unified information pipeline (Section~\ref{sec:pipeline}) and evaluate transfer under controlled stress tests that hold the detector fixed (Section~\ref{subsec:cross_domain_setup}), isolating domain effects from architectural confounds.
\textit{Validation evidence:} the Luderick stress test (Table~\ref{tab:scenario2_splitresults}) provides evidence that the fixed detector can retain usable performance under a habitat/appearance shift with relatively favourable visibility, supporting comparability claims within that operating regime.

\paragraph{(R2) Ecologists: False-negative asymmetry for early warning.}
For invasive or rare species monitoring, stakeholders prioritised avoiding missed detections over maximising aggregate accuracy.
\textit{Design response:} we treat recall (and recall-at-precision operating points) as a primary diagnostic and explicitly attribute false negatives to structural versus visual drivers under domain shift (Section~\ref{subsec:controlled_results}).
\textit{Validation evidence:} the DeepFish stress test (Table~\ref{tab:scenario1_splitresults}) exhibits a pronounced recall degradation in a low-redundancy regime, consistent with a sparsity-driven ``Context Collapse'' failure mode that is operationally critical for early warning.

\paragraph{(R3) Operators: Edge feasibility under constrained memory, power, and connectivity.}
Operators required that the system run on 4GB-class edge nodes with limited power budgets and intermittent connectivity, making continuous uplink impractical.
\textit{Design response:} these constraints bound model selection and motivate hardware-aware export and acceleration (Section~\ref{sec:edge_benchmark_setup}), evaluated on representative commodity hardware.
\textit{Validation evidence:} edge benchmarking demonstrates that TorchScript conversion can exceed a 4GB budget (OOM) while TensorRT enables practical sampling throughput (Table~\ref{tab:edge_benchmark}), establishing feasibility for long-horizon, low-rate monitoring policies.

\paragraph{(R4) Operators: Operational guardrails against spurious detections.}
Operators required guardrails against false alerts that inflate review workload and waste bandwidth in low-prevalence deployments.
\textit{Design response:} we retain background-only frames to preserve negative-class diversity and treat precision as an operational constraint rather than a secondary metric (Section~\ref{sec:pipeline}, Section~\ref{sec:results}).
\textit{Validation evidence:} baseline precision on the selected fixed detector (Table~\ref{tab:rf_results}) supports adherence to this guardrail and motivates its use in subsequent cross-domain analyses.

\section{Unified Information Pipeline}
\label{sec:pipeline}
We operationalise marine invasive species monitoring as an \emph{information reliability pipeline}: heterogeneous raw imagery is transformed into a standardised representation that supports controlled model training, comparable evaluation, and diagnostic analysis of failure modes under domain shift. 
A schematic of the end-to-end pipeline (dataset harmonisation, training, cross-domain tests, and edge benchmarking) is provided in Appendix~\ref{app:pipeline}.

\subsection{Heterogeneous Underwater Image Sources}
To support generalisation under long-term deployment variability, the pipeline integrates eight datasets summarised in Table~\ref{tab:datasets}. The sources span monitoring-relevant environments, from controlled aquaculture monocultures (AquaCoop) to complex, high-biodiversity reef scenes (e.g., Fish4Knowledge). They are selected to induce systematic variation along two axes: (i) \emph{scene structure} (density, overlap/occlusion, background dynamics) and (ii) \emph{visual quality} (turbidity/backscatter, blur/contrast, colour cast, illumination). This heterogeneity supports controlled cross-domain evaluation and enables diagnosis of whether reliability losses are primarily structural or visual. The integrated corpus contains 28{,}765 images and is split into 70\% training (20{,}130), 20\% validation (5{,}752), and 10\% testing (2{,}883) with the same class distribution. 

\begin{table}[t]
\centering
\resizebox{\columnwidth}{!}{%
\begin{tabular}{l l p{5.2cm}}
\toprule
\textbf{Dataset} & \textbf{Environment} & \textbf{Description \& key challenges (Structure / Visual)} \\
\midrule
\textbf{OzFish}~\cite{AIMS2019} & Coastal BRUVS &
High fish density and frequent overlap near bait (density/occlusion); variable illumination and backscatter (lighting/turbidity), with only 350 cleaned images retained from a larger BRUVS pool due to annotation errors.\\
\textbf{DeepFish}~\cite{Saleh2020} & Tropical reefs &
Sparse targets with low contextual redundancy (sparsity); strong blur and low contrast in turbid water (blur/contrast). \\
\textbf{Luderick}~\cite{GlobalWetlands2020} & Seagrass beds &
Dynamic vegetation and background motion (background dynamics); dominant green cast and visibility variation (colour cast/turbidity). \\
\textbf{AquaCoop} & Aquaculture cages &
Monoculture (\textit{D.\,labrax}) with net occlusion and overlapping silhouettes (occlusion/shape ambiguity); repetitive lighting patterns and specularities (illumination/reflections). \\
\textbf{Fish4Knowledge}~\cite{Fish4Knowledge2016} & Coral reefs &
Complex backgrounds and frequent small instances (background clutter/scale); high turbidity and colour attenuation (turbidity/colour). \\
\textbf{LifeCLEF}~\cite{Joly2015} & Tropical reefs &
High biodiversity with multiple co-occurring species (multi-object scenes); mixed illumination and viewpoint changes (lighting/viewpoint). \\
\textbf{Aquarium}~\cite{Dwyer2022} & Controlled tanks &
Lower background complexity with occasional glass artifacts (simple structure/reflections); artificial lighting and glare (illumination/specular). \\
\textbf{Open Ocean}~\cite{Roboflow2021} & Open ocean &
Wide variation in scene composition and scale (structure variability); strong natural-light variation (illumination). \\
\bottomrule
\end{tabular}%
}
\caption{Eight heterogeneous sources comprising the unified pipeline. The selection spans structural conditions (sparse vs.\ dense; low vs.\ high occlusion) and visual conditions (turbid vs.\ clear; colour cast), enabling controlled analysis of cross-domain reliability.}
\label{tab:datasets}
\end{table}

\subsection{Annotation Standardisation and Harmonisation}
\begin{figure}[t]
    \centering
    \includegraphics[width=\linewidth]{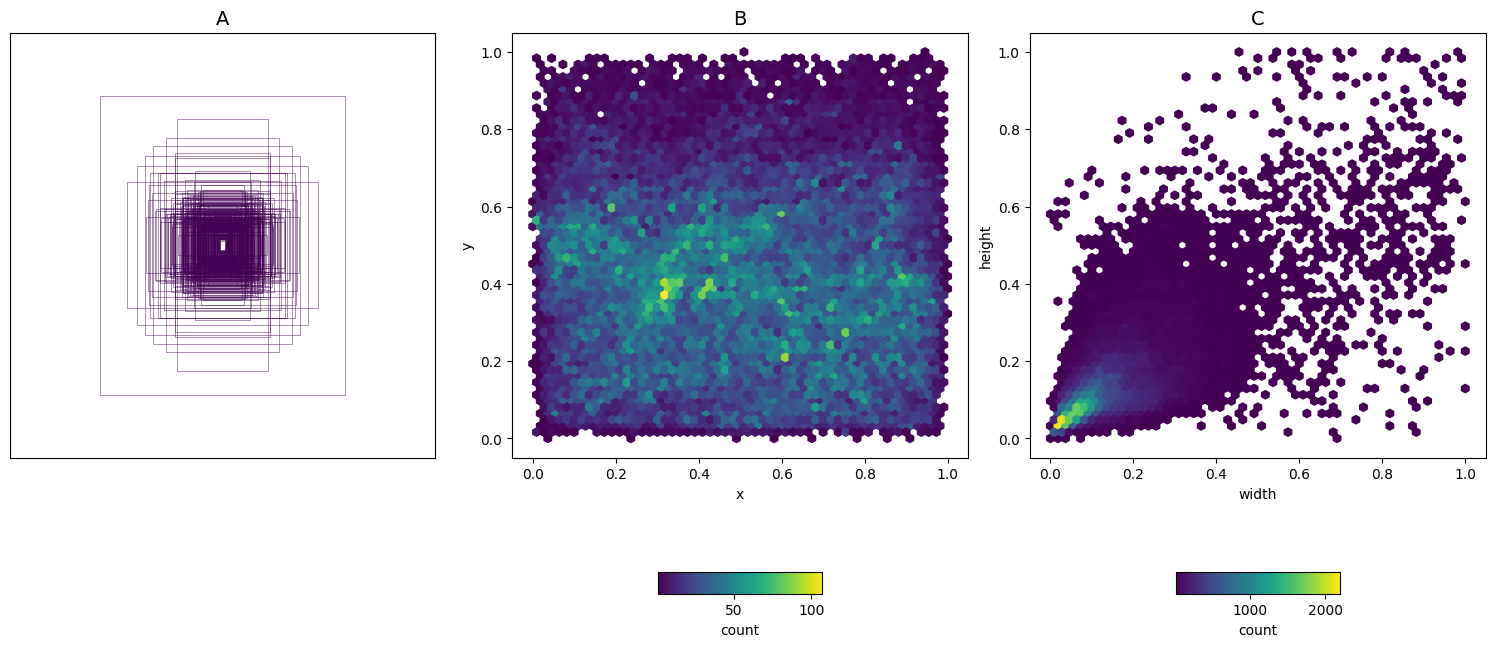}
    \caption{Unified annotation distribution. A) Spatial geometry of 300 sampled bounding boxes; B) Centroid density map; C) Width--height distribution illustrating variation in object size across datasets.}
    \label{fig:annotations_distribution}
\end{figure}

A prerequisite for cross-domain comparability is a unified annotation representation. All datasets are converted into a YOLO-compatible format where each instance is represented as \texttt{class\_id center\_x center\_y width height}, with all coordinates normalised to $[0,1]$ to ensure consistent geometry across images with different resolutions and aspect ratios. For datasets using polygon masks or XML contour formats, polygons are converted to minimal enclosing rectangles prior to normalisation and export. 
Figure~\ref{fig:annotations_distribution} summarises the unified annotation space, including sampled bounding boxes, centroid density, and width--height distributions.

\subsection{Data Integrity, Deduplication, and Split Protocol}
All images and labels undergo structured cleaning and validation prior to integration. To prevent leakage and reduce redundancy, we apply multi-stage deduplication: (i) path-level checks to resolve duplicate references, (ii) perceptual hashing (aHash) to detect near-duplicate frames, and (iii) MD5 hashing to detect exact duplicates. Hashes are computed for both images and label files; for each duplicate group, a single representative sample is retained.

To focus the study on detection reliability rather than taxonomy, all class labels are mapped to a single \texttt{fish} class. Images without valid labels after cleaning are retained as background-only samples to preserve negative-class diversity, reducing spurious detections in low-prevalence monitoring regimes.

The 70/20/10 split is constructed via a stratified two-step procedure. We first sample a held-out test set, then split the remaining data into training and validation. Stratification keys combine (a) source dataset identity and (b) presence/absence of fish annotations, ensuring balanced representation across datasets while preserving the positive/negative ratio in each subset. 

\subsection{Diagnostic Metrics: Visual Quality and Scene Structure}
\label{degradation_metrics}

Even under stratified splitting, substantial domain differences remain. To disentangle whether performance failures are driven primarily by visual degradation or by scene composition, we compute no-reference diagnostic metrics for every image and summarise them at the dataset level for descriptive analysis. These metrics are used only as covariates for interpretation, not for model optimisation.

\paragraph{Visual degradation metrics.}
We compute two established underwater image quality measures: UIQM~\cite{Panetta2015} and UCIQE~\cite{Yang2015}. To complement these perceptual scores, we compute additional proxies: (i) a haze/backscatter proxy (``turbidity'') based on image statistics, (ii) RGB channel ratios to characterise colour cast, (iii) RMS contrast, and (iv) a Laplacian-variance blur proxy.

\paragraph{Scene structure metrics.}
To test whether structural composition explains cross-domain failures beyond image quality, we compute: (v) Object Density, the number of annotated fish per image; and (vi) Structural Overlap, defined per image as the mean (over ground-truth boxes) of the maximum pairwise Intersection-over-Union (IoU) with any other ground-truth box. These metrics capture sparsity/clutter and occlusion/separability constraints that are not explained by visibility measures.

Collectively, these descriptors allow separation of domains that are visually degraded but structurally sparse from domains that are visually clean but crowded, supporting the controlled cross-domain analysis in subsequent sections. Due to page limitations, all metrics are detailed in Appendix~\ref{appendix:metrics}.

\section{Experimental Design for Cross-Domain Analysis}
\label{sec:experimental_design}

Using the unified pipeline, we hold the detection architecture fixed and vary the \emph{deployment domain} to quantify when extracted information remains comparable across environments. The design is structured to (i) select a stable, deployment-relevant detector, (ii) minimise confounding from suboptimal training, (iii) verify runtime feasibility on low-cost edge hardware, and (iv) test competing explanations for transfer failure: \emph{visual degradation} versus \emph{scene structure} using controlled stress tests grounded in the diagnostic metrics~\cite{hendrycks2019robustness}. Unless noted otherwise, we report standard detection metrics (mAP, precision, recall) to characterise both accuracy and failure asymmetries relevant to monitoring.

\subsection{Detection Model Selection}
\label{subsec:model_selection}

We benchmark four YOLO variants under identical training conditions to select a baseline that is both (a) stable for controlled cross-domain analysis and (b) feasible under field deployment constraints. The evaluated models are YOLOv8m~\cite{varghese2024yolov8,li2025efficient,mohankumar2025benchmark} and YOLO11n/YOLO11s/YOLO11m~\cite{khanam2024yolov11}. Each model is trained with the same schedule and evaluated on the Roboflow Aquarium baseline~\cite{Dwyer2022}, with results reported in Table~\ref{tab:rf_results}.

We select \textbf{YOLO11m} as the baseline~\cite{liu2025lfn,vijayalakshmi2025aquayolo,wilson2025automated}. The selection is motivated by output stability under repeated runs, its practical compute/memory profile relative to larger variants that are unlikely to fit common edge devices, and reduced deprecation risk given its recent release, as older architectures may lose framework support over the multi-year deployment timeline. We additionally follow Ultralytics guidance that prioritises YOLO11 for predictable memory usage and stable training behaviour in deployment-oriented settings~\cite{UltralyticsYOLO12}. Importantly, the goal of this step is not to claim state-of-the-art accuracy, but to anchor subsequent cross-domain analyses on a single, reproducible detector.

\subsection{Training and Optimisation Protocol}
\label{subsec:training_optim}

To ensure that observed performance differences are attributable to domain shift rather than training instability, YOLO11m is trained with a fixed, \emph{pre-specified} optimisation protocol. Hyperparameters are selected via Bayesian optimisation using Optuna~\cite{akiba2019optunanextgenerationhyperparameteroptimization}, motivated by sample-efficient search in high-dimensional spaces~\cite{9451544,yu2020hyperparameteroptimizationreviewalgorithms}. Optimisation proceeds in two phases:

\paragraph{Phase 1: Core training parameters.}
We optimised the choice of optimiser and learning-rate schedule, regularisation parameters (e.g., dropout), and loss weights, with particular emphasis on bounding-box localisation to support small-object detection. 

\paragraph{Phase 2: Augmentation parameters.}
With the core configuration fixed, we optimise augmentation strength~\cite{liu2024cross}. The final policy uses strong mosaic augmentation to increase exposure to occlusions and multi-object scenes, while limiting aggressive colour transformations. This constraint is deliberate: excessive colour jitter can inflate false positives in empty-water frames, which is operationally undesirable in monitoring settings where spurious detections can distort abundance indices and trigger unnecessary reviews. 

\subsection{Operational Feasibility: Edge Deployment Benchmarking}
\label{sec:edge_benchmark_setup}

Because monitoring deployments often operate under limited power, memory, and connectivity, we evaluate whether the selected detector can sustain near-continuous inference on low-cost edge hardware. We target the NVIDIA Jetson Nano (4GB), representative of field-deployable nodes in remote or resource-limited settings. We measure inference latency (ms) and throughput (FPS) under three deployment formats: i) PyTorch (native): development baseline;
ii) TorchScript: portable serialisation for embedded inference; and iii) TensorRT (FP16): hardware-optimised inference with kernel fusion and reduced precision. This protocol quantifies the performance gap between research execution and field constraints, and supports later discussion of economically scalable monitoring configurations.

\subsection{Cross-Domain Evaluation Setup}
\label{subsec:cross_domain_setup}

We evaluate transfer reliability using two complementary components.

\paragraph{Unseen-domain evaluation.}
The trained model is applied to external datasets not used during training or optimisation to probe behaviour beyond the training distribution. The Fish Video Object Tracking dataset~\cite{kaggle_fish_object_tracking} contains synthetic underwater scenes with stylised textures and comparatively uniform backgrounds, while the Kaggle Fish Detection dataset~\cite{kaggle_fish_detection} contains dense aquaculture footage with frequent occlusion and high object density. These datasets provide additional evidence that performance changes are not artefacts of the unified split construction.

\paragraph{Controlled stress tests informed by diagnostic metrics.}
To explicitly test whether cross-domain performance loss is driven primarily by \emph{visual degradation} or by \emph{scene structure}, we compute the diagnostic metrics of Section~\ref{degradation_metrics} across all domains and construct two targeted test splits. Each split isolates a distinct failure hypothesis while holding the detector fixed:

\begin{itemize}
  \setlength{\itemsep}{0pt}
  \setlength{\parskip}{0pt}
  \setlength{\parsep}{0pt}
    \item \textbf{Scenario 1 (DeepFish): Visual degradation with structural sparsity.} The test split contains only \textbf{DeepFish}~\cite{Saleh2020}, a visually degraded domain (strong blur, low contrast) that is also structurally sparse (low object density and limited contextual redundancy). This scenario probes whether poor visibility alone explains transfer failure, or whether sparse structure causes disproportionate degradation even when targets are visually discernible.
    \item \textbf{Scenario 2 (Luderick): Habitat/colour shift with dynamic structure.} The test split contains only \textbf{Luderick}~\cite{GlobalWetlands2020}, which is comparatively cleaner in visibility but exhibits a strong habitat and colour shift (green seagrass background) and dynamic backgrounds. This scenario probes sensitivity to background semantics and structural dynamics under relatively favourable visual quality.
\end{itemize}

Together, these stress tests link performance variation to measurable domain properties (visual quality and structure metrics) rather than to architectural changes. This allows to attribute observed failures to specific, operationally interpretable drivers (e.g., sparsity versus turbidity; occlusion versus colour cast), supporting stakeholder-facing conclusions about when automated outputs can be treated as comparable monitoring evidence~\cite{hendrycks2019robustness}.

\section{Results}
\label{sec:results}

We evaluate the unified pipeline along three axes required for monitoring-grade information extraction: (i) baseline detection performance and model stability, (ii) transfer reliability under domain shift, and (iii) operational feasibility on low-cost edge hardware.

\subsection{Baseline Model Selection}
Table~\ref{tab:rf_results} compares four YOLO variants trained under identical conditions on the Roboflow Aquarium baseline. YOLOv8m achieves the strongest overall performance in mAP50--95, recall, and F1-score, while YOLO11m yields the highest precision and mAP50 with competitive overall accuracy.

For monitoring, precision is a critical guardrail against false positives that can bias downstream abundance indicators and inflate manual review effort. Together with its stable training behaviour and predictable memory profile in deployment-oriented settings, these considerations motivate selecting YOLO11m as the fixed baseline for subsequent analyses.

\begin{table}[h]
    \centering
    \resizebox{\columnwidth}{!}{%
    \begin{tabular}{l r r r r r r}
        \toprule
        \textbf{Model} & \textbf{Params (M)} & \textbf{mAP50--95} & \textbf{mAP50} & \textbf{Precision} & \textbf{Recall} & \textbf{F1} \\
        \midrule
        YOLOv8m & 25.9 & \textbf{0.4846} & \underline{0.7901} & \underline{0.7837} & \textbf{0.7356} & \textbf{0.7589} \\
        YOLO11n & 2.6  & 0.3595 & 0.6977 & 0.7255 & 0.6102 & 0.6629 \\
        YOLO11s & 9.4  & \underline{0.4703} & 0.7499 & 0.7763 & 0.6588 & 0.7127 \\
        YOLO11m & 20.1 & 0.4657 & \textbf{0.8097} & \textbf{0.8117} & \underline{0.6866} & \underline{0.7439} \\
        \bottomrule
    \end{tabular}%
    }
    \caption{Comparison of YOLO models on the Roboflow Aquarium dataset. YOLO11m is selected for its superior precision.}
    \label{tab:rf_results}
\end{table}

\subsection{In-Domain Optimisation and Resolution}
YOLO11m was fine-tuned at two resolutions ($640\times640$ and $1024\times1024$) to test whether higher resolution yields meaningful gains. Increasing resolution to 1024 pixels yielded negligible changes in mAP and F1 (Table~\ref{tab:finetuning_resolution_comparison}). Therefore, the $640\times640$ configuration is retained as the operational baseline, preserving efficiency without sacrificing reliability.

\begin{table}[h]
    \centering
    \resizebox{\columnwidth}{!}{%
    \begin{tabular}{lrrrrrr}
    \toprule
    \textbf{Configuration} & \textbf{mAP50--95} & \textbf{mAP50} & \textbf{Precision} & \textbf{Recall} & \textbf{F1} \\
    \midrule
    Validation (640px)  & \textbf{0.624} & \textbf{0.918} & \textbf{0.910} & \textbf{0.867} & \textbf{0.889} \\
    Validation (1024px) & 0.621 & 0.913 & 0.907 & 0.866 & 0.886 \\
    Test (640px)        & \textbf{0.615} & \textbf{0.917} & \textbf{0.912} & \textbf{0.866} & \textbf{0.888} \\
    Test (1024px)       & 0.612 & 0.912 & 0.911 & 0.858 & 0.884 \\
    \bottomrule
    \end{tabular}%
    }
    \caption{In-domain performance of YOLO11m. Higher resolution offers diminishing returns.}
    \label{tab:finetuning_resolution_comparison}
\end{table}

\subsection{Generalisation to External Domains}
To probe behaviour beyond the unified split, the model was evaluated on two external datasets (Table~\ref{tab:unseen_datasets_comparison}). Performance remains strong on synthetic scenes (high precision), indicating robustness to texture shifts. In contrast, performance degrades sharply on dense aquaculture footage, where frequent occlusion reduces separability (Recall 0.318). This motivates the controlled experiments below.

\begin{table}[h]
    \centering
    \resizebox{\columnwidth}{!}{%
    \begin{tabular}{lrrrrr}
        \toprule
        \textbf{Dataset} & \textbf{Images} & \textbf{mAP50-95} & \textbf{Precision} & \textbf{Recall} & \textbf{F1} \\
        \midrule
        Synthetic (Tracking) & 99  & 0.557 & 0.827 & 0.715 & 0.767 \\
        Aquaculture (Dense)  & 629 & 0.255 & 0.804 & 0.318 & 0.456 \\
        \bottomrule
    \end{tabular}%
    }
    \caption{Performance on external unseen datasets. High density drives failure more than synthetic texture shifts.}
    \label{tab:unseen_datasets_comparison}
\end{table}

\subsection{Controlled Cross-Domain Stress Tests}
\label{subsec:controlled_results}
We conducted two leave-one-domain-out stress tests. DeepFish and Luderick were selected as target domains to represent distinct failure hypotheses (Table~\ref{tab:distortion_summary}).

\begin{table*}[t]
    \centering
    \resizebox{\linewidth}{!}{%
    \begin{tabular}{lrrrrrrrrrrr}
        \toprule
        \textbf{Dataset} & \textbf{Turbidity} & \textbf{Contrast} & \textbf{BlurVar} & \textbf{Blue} & \textbf{Green} & \textbf{Red} & \textbf{UIQM} & \textbf{UCIQE} & \textbf{FishCount} & \textbf{Overlap} & \textbf{Difficulty} \\
        \midrule
        OzFish     & 0.00 & 0.80 & 0.88 & \textbf{1.00} & \textbf{1.00} & \textbf{1.00} & 0.48 & \textbf{1.00} & \textbf{1.00} & \textbf{1.00} & \textbf{0.82} \\
        DeepFish   & \textbf{0.95} & \underline{0.81} & \underline{0.98} & 0.31 & 0.37 & 0.33 & \underline{0.57} & \textbf{1.00} & 0.16 & 0.24 & \underline{0.57} \\
        Luderick   & 0.32 & 0.64 & 0.75 & 0.49 & \underline{0.99} & \underline{0.71} & 0.29 & \textbf{1.00} & 0.10 & 0.13 & 0.54 \\
        AquaCoop   & 0.33 & \textbf{1.00} & \textbf{1.00} & 0.25 & 0.00 & 0.12 & \textbf{1.00} & 0.00 & \underline{0.51} & \underline{0.97} & 0.52 \\
        \bottomrule
    \end{tabular}%
    }
    \caption{Normalised distortion and structure metrics. DeepFish represents Visual Degradation; OzFish represents Structural Complexity.}
    \label{tab:distortion_summary}
\end{table*}

\subsubsection{Scenario 1: Context Collapse (DeepFish)}
\begin{figure}[h]
    \centering
    \includegraphics[width=\linewidth]{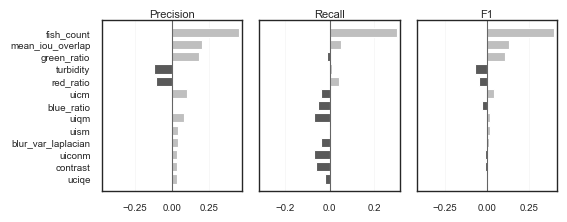}
    \caption{Horizontal Correlation Bars for Scenario 1. Performance tracks Structure (FishCount) more than Vision (Turbidity).}
    \label{fig:Senario1_corrbars}
\end{figure}

When DeepFish is excluded, performance drops sharply (Table~\ref{tab:scenario1_splitresults}). Correlation analysis (Figure~\ref{fig:Senario1_corrbars}) reveals this is driven by structural sparsity, not visual quality. DeepFish scenes lack contextual redundancy (schooling, habitat complexity); without these cues, the detector struggles to verify isolated targets, leading to a collapse in recall.

\begin{table}[h]
    \centering
    \resizebox{\columnwidth}{!}{%
    \begin{tabular}{lrrrr}
        \toprule
        \textbf{Split} & \textbf{mAP50--95} & \textbf{Precision} & \textbf{Recall} & \textbf{F1} \\
        \midrule
        Validation & 0.604 & 0.891 & 0.835 & 0.862 \\
        Test (DeepFish) & 0.285 & 0.616 & \textbf{0.552} & 0.583 \\
        \bottomrule
    \end{tabular}
    }
    \caption{Scenario 1: Testing on DeepFish. Recall drops due to sparsity.}
    \label{tab:scenario1_splitresults}
\end{table}

\begin{figure}[h]
    \centering
    \includegraphics[width=0.8\linewidth]{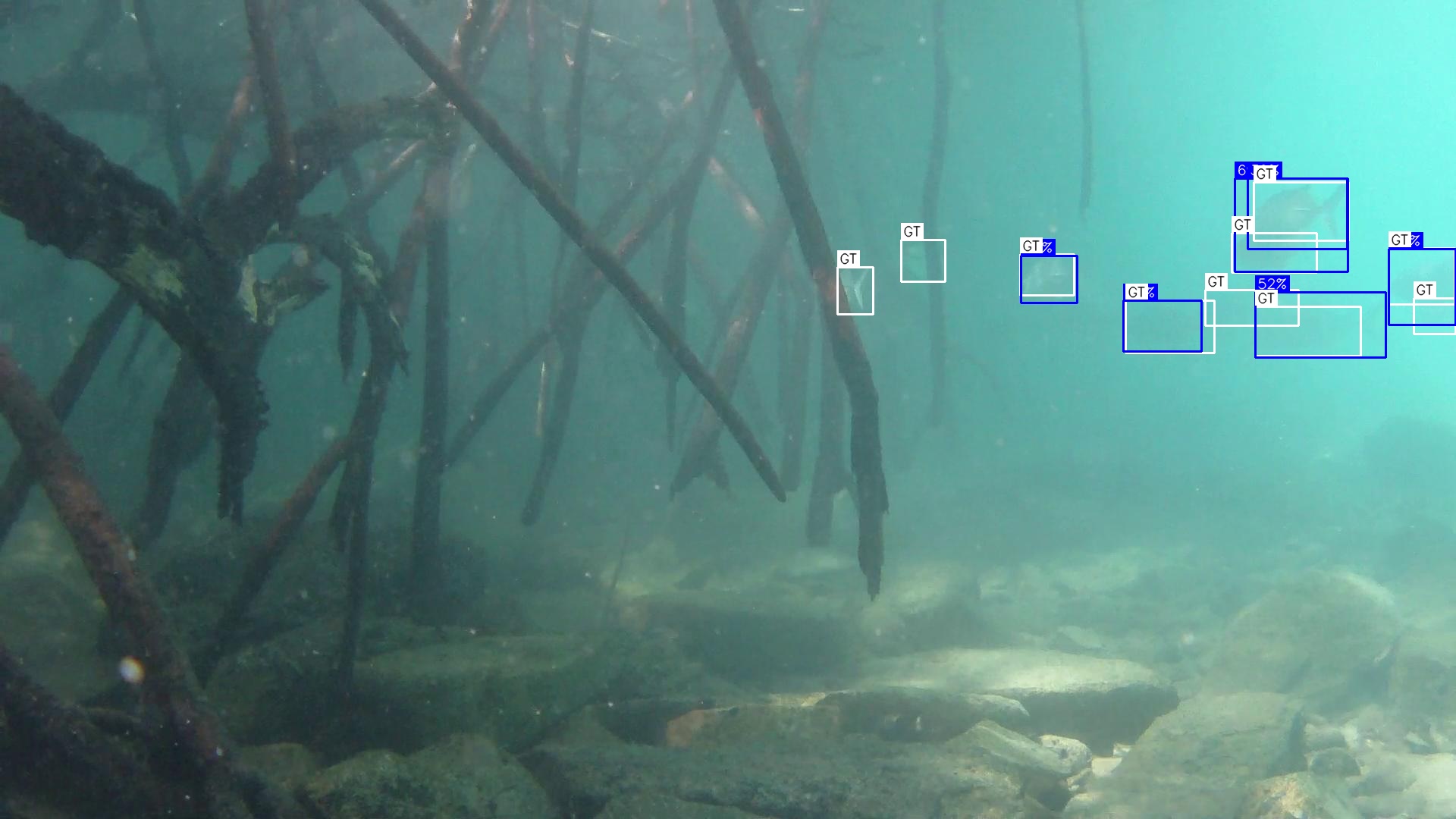}
    \caption{Sample inference image from scenario 1 (DeepFish). Blue bounding boxes indicate predictions, while white boxes denote ground truth.}
    \label{fig:scenario1_test}
\end{figure}

\subsubsection{Scenario 2: Semantic Shift (Luderick)}
When Luderick is excluded, the performance drop is moderate (Table~\ref{tab:scenario2_splitresults}). Despite strong colour/habitat shifts, the model generalises better than in DeepFish, confirming that visual/semantic shifts are less destabilising than structural ones.

\begin{figure}[h]
    \centering
    \includegraphics[width=0.6\linewidth]{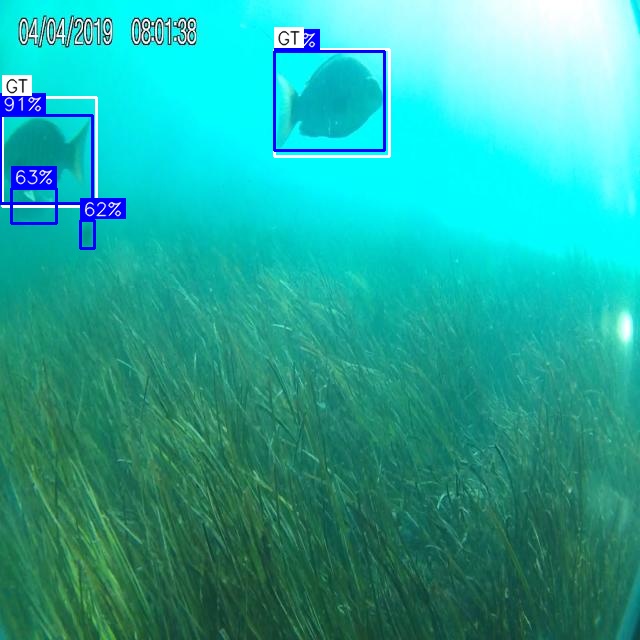}
    \caption{Sample inference image from scenario 2 (Luderick). Blue bounding boxes indicate predictions, while white boxes denote ground truth.}
    \label{fig:scenario2_test}
\end{figure}

\begin{table}[h]
    \centering
    \resizebox{\columnwidth}{!}{%
    \begin{tabular}{lrrrr}
        \toprule
        \textbf{Split} & \textbf{mAP50--95} & \textbf{Precision} & \textbf{Recall} & \textbf{F1} \\
        \midrule
        Validation & 0.608 & 0.899 & 0.850 & 0.874 \\
        Test (Luderick) & 0.497 & 0.752 & 0.698 & 0.724 \\
        \bottomrule
    \end{tabular}
    }
    \caption{Scenario 2: Testing on Luderick. Robustness is retained despite domain shift.}
    \label{tab:scenario2_splitresults}
\end{table}

\begin{figure}[h]
    \centering
    \includegraphics[width=\linewidth]{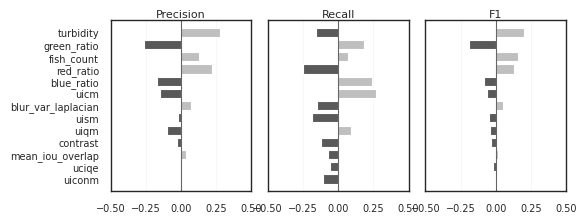}
    \caption{Horizontal Correlation Bars for Scenario 2.}
    \label{fig:Senario2_corrbars}
\end{figure}

\subsubsection{Attribution of Failure Drivers: Structure vs. Vision}
\label{subsec:attribution}

To attribute failure drivers without relying on small-sample domain correlations, we stratify evaluation frames by (i) \emph{object density} (fish per image; bins: 0, 1, 2--3, 4--7, $\geq 8$) and (ii) \emph{structural overlap} (mean max-IoU among ground-truth boxes). Within each stratum, we compute recall and compare trends against stratification by visual covariates (turbidity, blur).

Across domains, recall is lowest in the sparsest strata, and the largest recall gaps occur between low-density and higher-density bins, consistent with the "Context Collapse'' failure mode. In contrast, stratification by visual degradation yields weaker and less consistent gradients, indicating that visibility alone is not the dominant determinant of transfer reliability.

\subsection{Operational Feasibility: Edge Deployment}
\label{sec:edge_results}
We benchmarked inference on an \textbf{NVIDIA Jetson Nano (4GB)}, representative of commodity low-cost field nodes ($<\$150$). Benchmarks were executed at $640\times640$ resolution with batch size 1 to reflect deployment latency constraints (Table~\ref{tab:edge_benchmark}).

\paragraph{Engineering constraints.}
Under these settings, TorchScript conversion for YOLO11m exceeded the 4GB memory budget (\textit{OOM}). This failure mode, invisible on server-grade hardware, highlights the necessity of hardware-aware optimisation in field deployments.

\paragraph{Monitoring policy feasibility.}
TensorRT (FP16) yields \textbf{3.45 FPS}. While insufficient for high-frequency tracking, this throughput supports periodic sampling policies used to approximate frame-based abundance indices (e.g., \emph{MaxN} computed over fixed time windows), enabling continuous low-rate sampling or burst-mode regimes that prioritise long-horizon coverage over per-second trajectory fidelity.

\begin{table}[h]
    \centering
    \resizebox{\columnwidth}{!}{%
    \begin{tabular}{l l r r r}
        \toprule
        \textbf{Model} & \textbf{Format} & \textbf{Latency} & \textbf{FPS} & \textbf{Gain} \\
        \midrule
        YOLO11n & PyTorch    & 111ms & 8.99 & -- \\
        YOLO11n & TensorRT   & 88ms  & 11.27 & $+25\%$ \\
        \midrule
        YOLO11m & PyTorch    & 450ms & 2.22 & -- \\
        YOLO11m & TorchScript & \multicolumn{3}{c}{\textit{Failed (OOM)}} \\
        YOLO11m & TensorRT   & 290ms & 3.45 & $+55\%$ \\
        \bottomrule
    \end{tabular}%
    }
    \caption{Edge inference on Jetson Nano. TensorRT makes Medium-sized models feasible for sampling.}
    \label{tab:edge_benchmark}
\end{table}

\section{Discussion and Implications}
\label{sec:discussion}

This study reframes underwater fish detection from maximising in-domain benchmark scores to producing \emph{monitoring-grade information} that remains comparable under domain shift and feasible under field constraints. Holding the detector fixed and varying deployment domains shows that transfer failures align more consistently with \emph{scene structure} (sparsity, density, overlap) than with visual quality measures alone. Practically, this implies that improving visibility via enhancement is, by itself, an incomplete remedy: reliability depends on whether the scene provides sufficient contextual redundancy for verification and sufficient separability under occlusion.

\paragraph{Implications for monitoring protocols and data design.}
For practitioners, the actionable lever is \emph{structural coverage}. Acquisition and curation should explicitly include (i) sparse regimes with isolated targets and (ii) crowded regimes with overlap/occlusion, rather than optimising only for clearer imagery. Reporting performance by interpretable strata (e.g., density and overlap bins) provides a deployment-facing reliability statement that is more portable than dataset names or single-number mAP summaries, and it supports cross-site comparability by linking expected error modes to observable site conditions.

\paragraph{Operational guardrails and reporting.}
Because false alerts impose review and bandwidth costs in low-prevalence deployments, precision should be treated as a guardrail and recall should be interpreted at operating points consistent with that guardrail (e.g., recall-at-fixed-precision). Under this framing, “reliability” is not a global property of the detector but a qualified claim conditional on structural regime (sparse versus crowded) and the operational threshold.

\paragraph{Edge feasibility and time-to-information.}
Edge benchmarking shows that deployment viability depends on the export/acceleration path. On 4GB-class nodes, naive serialisation may fail (OOM), whereas TensorRT enables practical throughput for periodic sampling policies. At 3.45 FPS, processing 1{,}000 frames is on the order of minutes (approximately 4.8 minutes), supporting long-horizon, low-rate monitoring without continuous uplink.

\subsection{Limitations}
\label{subsec:limitations}

\paragraph{Scope of claims.}
The analysis is intentionally architecture-controlled (a fixed detector) to support attribution under domain shift. Findings should be interpreted as evidence about \emph{data- and environment-driven} reliability limits, not as a statement of detector state-of-the-art.

\paragraph{Binary detection abstraction.}
Collapsing taxonomy to a single \texttt{fish} class establishes a robust detection foundation but does not resolve species-level confusion, which is central for invasive-species decisions. Structure-driven failures may interact differently with fine-grained classification, particularly for rare taxa and small objects.

\paragraph{Frame-level evidence.}
The pipeline produces frame-level detections suitable for relative abundance proxies (e.g., MaxN-style indices) and alerting, but it does not recover identity-level continuity in dense shoals. Occlusion and group motion therefore remain constraints for behavioural inference and absolute counting.

\paragraph{Coverage of extreme regimes.}
Although the corpus is heterogeneous, extreme conditions (severe biofouling, near-blackout turbidity, very low-light) are underrepresented. Reliability in these regimes likely requires targeted data acquisition and human-in-the-loop fallback.

\subsection{Future Work}
\label{subsec:futurework}

Future work should (i) incorporate temporal consistency via lightweight tracking to mitigate transient misses, (ii) curate training data by \emph{structural strata} (density/overlap) rather than dataset identity alone, (iii) add human-in-the-loop active learning that prioritises structurally out-of-distribution frames (especially sparse regimes associated with recall collapse), and (iv) evaluate tiered edge--cloud monitoring where the edge filters routine frames and transmits ambiguous cases for higher-capacity verification.

\section{Conclusion}
\label{sec:conclusion}

This work shows that reliable marine monitoring requires more than strong in-domain detection accuracy. By implementing a Unified Information Pipeline and evaluating a fixed detector under controlled cross-domain stress tests, we find that performance loss under transfer is explained more by \emph{scene structure} (specifically, the ``Context Collapse" phenomenon in sparse domains) than by visual degradation metrics alone. This challenges common assumption that image enhancement is the primary remedy for underwater AI~\cite{awad2024beneath,liu2023image}, shifting emphasis toward structure-aware diagnostics as a prerequisite for comparable monitoring evidence.

Crucially, the pipeline is operationally grounded within an ongoing, multi-year European marine conservation initiative targeting invasive species management in Arctic and Atlantic waters. Edge benchmarking demonstrates that deployment-relevant inference is achievable on low-cost, resource-constrained hardware suitable for remote northern installations, supporting economically scalable monitoring in settings with limited connectivity. This foundational detection stage establishes reliability requirements for extending to species-level classification in operational deployments. Overall, the proposed system functions as an information-support tool: it standardises heterogeneous imagery, clarifies when automated outputs can be trusted across domains, and provides the technical foundation for experts to focus on the ecological interpretation required to protect marine biodiversity.


\clearpage
\bibliographystyle{named}
\bibliography{ijcai26}

\clearpage
\appendix


\clearpage
\section{Metrics}
\label{appendix:metrics}

\subsection{Model Performance Metrics}
Intersection over Union (IoU) measures the overlap between a predicted bounding box and its corresponding ground-truth annotation. It is the fundamental criterion behind all mAP computation. IoU is defined as:
\[
\text{IoU} = \frac{|B_{\text{pred}} \cap B_{\text{gt}}|}{|B_{\text{pred}} \cup B_{\text{gt}}|}
\]
\\

\paragraph{mAP50}
The mean Average Precision at IoU = 0.50 (mAP50) evaluates whether the model can correctly 
identify and localise fish under a relatively permissive bounding-box overlap requirement:
\[
\text{mAP50} = \frac{1}{N} \sum_{i=1}^{N} AP_i(\text{IoU}=0.50)
\]
where $AP_i$ denotes the Average Precision for class $i$. In this study, the task is binary 
(fish vs. background), so $N = 1$.\\

\paragraph{mAP50-95}
The mean Average Precision computed across ten IoU thresholds from 0.50 to 0.95 in steps of 0.05, following the COCO evaluation protocol. This metric imposes 
increasingly strict localisation requirements and provides a more comprehensive view of the detector's performance:
\[
\text{mAP50-95} = \frac{1}{10} \sum_{t = 0.50}^{0.95} AP(\text{IoU} = t)
\]
\\
\paragraph{Precision}
Precision quantifies the proportion of predicted positive detections that are correct. High precision indicates that the model produces few false positives, which is important in underwater scenes where background clutter or floating particles may cause spurious detections:
\[
\text{Precision} = \frac{TP}{TP + FP}
\]
\\
\paragraph{Recall}
Recall reflects how many true objects in the image received at least one correct bounding-box prediction. A high recall value indicates that the detector can capture most fish instances, even under challenging conditions such as turbidity, occlusion, low contrast, or motion blur:
\[
\text{Recall} = \frac{TP}{TP + FN}
\]

\paragraph{F1-Score}
The F1-score is the harmonic mean of Precision and Recall. This makes it particularly appropriate for underwater datasets, which often contain class imbalance:
\[
\text{F1} = 2 \cdot \frac{\text{Precision} \cdot \text{Recall}}{\text{Precision} + \text{Recall}}
\]
\\


\subsection{Image Degradation and Quality Metrics}

\paragraph{Turbidity}
Turbidity estimates the level of haze caused by suspended particles in water. It is computed using a dark-channel prior proxy, where higher values indicate stronger scattering and reduced visibility:
\[
\text{Turbidity} = \frac{1}{|\Omega|} \sum_{x \in \Omega} \min_{c \in \{r,g,b\}} I^c(x)
\]
where $\Omega$ denotes the image domain and $I^c(x)$ is the intensity of colour channel $c$ at pixel $x$.\\

\paragraph{RMS Contrast}
RMS contrast measures luminance variability in the image and reflects global visibility conditions:
\[
\text{Contrast}_{\text{RMS}} =
\sqrt{\frac{1}{|\Omega|} \sum_{x \in \Omega} \left(I_L(x) - \mu_L\right)^2}
\]
\\

\paragraph{Blur (Laplacian Variance)}
Image blur is estimated using the variance of the Laplacian operator, which captures edge strength:
\[
\text{Blur}_{\text{var}} = \operatorname{Var}\!\left(\nabla^2 I_L\right)
\]
\\


\subsection{Colour Distribution Metrics}

\paragraph{Red, Green, and Blue Ratios}
Colour ratios quantify channel dominance and underwater colour attenuation effects:
\[
R_c = \frac{\sum_{x \in \Omega} I^c(x)}
{\sum_{x \in \Omega} \left(I^r(x)+I^g(x)+I^b(x)\right)}, \quad c \in \{r,g,b\}
\]
with $R_r + R_g + R_b = 1$.\\


\subsection{Underwater Image Quality Metrics}

\paragraph{UIQM}
The Underwater Image Quality Measure (UIQM) combines colourfulness, sharpness, and contrast into a single score:
\[
\text{UIQM} = c_1 \cdot \text{UICM} + c_2 \cdot \text{UISM} + c_3 \cdot \text{UIConM}
\]
where $c_1 = 0.0282$, $c_2 = 0.2953$, and $c_3 = 3.5753$.\\

\begin{itemize}
  \setlength{\itemsep}{0pt}
  \setlength{\parskip}{0pt}
  \setlength{\parsep}{0pt}
    \item \textbf{UICM} The Underwater Image Colourfulness Measure evaluates colour balance and distortion:
\[
\text{UICM} =
- \sqrt{\mu_{RG}^2 + \mu_{YB}^2}
- 0.3 \sqrt{\sigma_{RG}^2 + \sigma_{YB}^2}
\]
where $RG = R - G$ and $YB = \frac{R+G}{2} - B$.\\
    \item \textbf{UISM} The Underwater Image Sharpness Measure quantifies edge strength using gradient magnitude:
\[
\text{UISM} =
\frac{1}{|\Omega|}
\sum_{x \in \Omega}
\sqrt{G_x(x)^2 + G_y(x)^2}
\]
where $G_x$ and $G_y$ are Sobel gradients.\\

    \item \textbf{UIConM} The Underwater Image Contrast Measure evaluates local contrast using Michelson contrast over image blocks:
\[
\text{UIConM} =
\frac{1}{N}
\sum_{i=1}^{N}
\frac{I_{\max}^{(i)} - I_{\min}^{(i)}}
{I_{\max}^{(i)} + I_{\min}^{(i)}}
\]
\end{itemize}

\paragraph{UCIQE}
The Underwater Colour Image Quality Evaluation metric focuses on chroma dispersion, luminance contrast, and saturation:
\[
\text{UCIQE} =
\alpha \sigma_c + \beta \, \text{con}_L + \gamma \mu_s
\]
with $\alpha = 0.4680$, $\beta = 0.2745$, and $\gamma = 0.2576$.\\


\subsection{Scene Structure and Complexity Metrics}

\paragraph{Fish Count}
Fish count measures scene density as the number of annotated objects in an image:
\[
\text{FishCount} = |B|
\]
where $B$ is the set of ground-truth bounding boxes.\\

\paragraph{Mean Bounding Box Overlap}
Mean overlap quantifies occlusion severity using the average pairwise Intersection over Union between bounding boxes:
\[
\text{MeanOverlap} =
\frac{2}{N(N-1)}
\sum_{i<j}
\frac{|B_i \cap B_j|}{|B_i \cup B_j|}
\]
where $N$ is the number of bounding boxes in the image.

\section{Pipeline Schematic}
\label{app:pipeline}

This schematic provides a compact view of the full workflow: dataset aggregation and harmonisation, fixed-detector training with Optuna-tuned hyperparameters, external-domain checks, controlled stress tests, and edge optimisation/deployment.

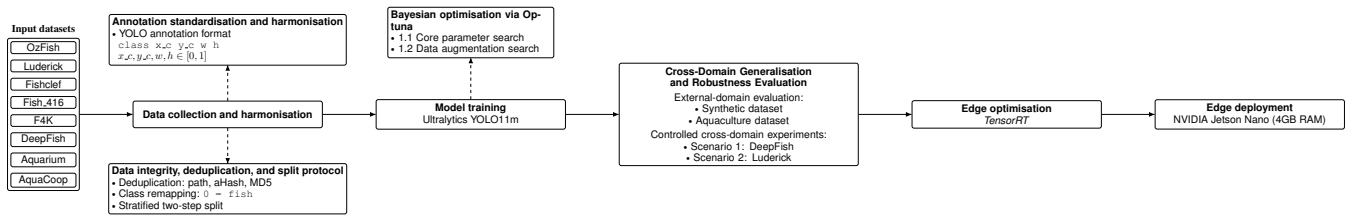
\begin{figure*}[b!]
\centering
\resizebox{\textwidth}{!}{%
\begin{tikzpicture}[
  font=\sffamily\normalsize,
  node distance=10mm and 16mm
]

\node[datasetBox] (d1) {OzFish};
\node[datasetBox, below=2mm of d1] (d2) {Luderick};
\node[datasetBox, below=2mm of d2] (d3) {Fishclef};
\node[datasetBox, below=2mm of d3] (d4) {Fish\_416};
\node[datasetBox, below=2mm of d4] (d5) {F4K};
\node[datasetBox, below=2mm of d5] (d6) {DeepFish};
\node[datasetBox, below=2mm of d6] (d7) {Aquarium};
\node[datasetBox, below=2mm of d7] (d8) {AquaCoop};

\node[group, fit=(d1)(d8),
      label={[title]north:Input datasets}] (dsgrp) {};

\node[box, right=18mm of dsgrp] (harm)
{\textbf{Data collection and harmonisation}};

\node[boxAlt, right=18mm of harm] (train)
{\textbf{Model training}\\Ultralytics YOLO11m};

\node[box, right=18mm of train, text width=78mm] (crossdomain) {
  \textbf{Cross-Domain Generalisation}\\
  \textbf{and Robustness Evaluation}\\[3pt]
  External-domain evaluation:\\
  \bb Synthetic dataset\\
  \bb Aquaculture dataset\\[3pt]
  Controlled cross-domain experiments:\\
  \bb Scenario 1: DeepFish\\
  \bb Scenario 2: Luderick
};

\node[boxAlt, right=18mm of crossdomain] (export)
{\textbf{Edge optimisation}\\\textit{TensorRT}};

\node[box, right=18mm of export] (edge)
{\textbf{Edge deployment}\\NVIDIA Jetson Nano (4GB RAM)};

\node[note, above=12mm of harm, anchor=south] (noteL) {
  \textbf{Annotation standardisation and harmonisation}\\
  \bb YOLO annotation format\\
  \hspace{2mm}\texttt{class x\_c y\_c w h}\\
  \hspace{2mm}$x\_c, y\_c, w, h \in [0,1]$
};

\node[note, below=12mm of harm, anchor=north] (noteR) {
  \textbf{Data integrity, deduplication, and split protocol}\\
  \bb Deduplication: path, aHash, MD5\\
  \bb Class remapping: \texttt{0 = fish}\\
  \bb Stratified two-step split
};

\node[note, above=14mm of train, anchor=south, text width=54mm] (optuna) {
  \textbf{Bayesian optimisation via Optuna}\\
  \bb 1.1 Core parameter search\\
  \bb 1.2 Data augmentation search
};

\draw[arrow] (dsgrp.east) -- (harm.west);
\draw[arrow] (harm.east) -- (train.west);
\draw[arrow] (train.east) -- (crossdomain.west);
\draw[arrow] (crossdomain.east) -- (export.west);
\draw[arrow] (export.east) -- (edge.west);

\draw[dashedarrow] (harm.north) -- (noteL.south);
\draw[dashedarrow] (harm.south) -- (noteR.north);
\draw[dashedarrow] (train.north) -- (optuna.south);

\end{tikzpicture}%
}

\caption{Pipeline from heterogeneous underwater datasets to a single-class fish detector (YOLO11m), including cross-domain evaluation and robustness analysis, and edge deployment on NVIDIA Jetson Nano.}
\label{fig:pipeline}
\end{figure*}

\end{document}